\documentclass{llncs}
\usepackage{makeidx}  

\usepackage{hyperref}				
\hypersetup{backref,colorlinks=true}	
\usepackage{times}              		
\usepackage{graphicx}           		
\usepackage{subfigure}          		
\usepackage{tabularx}		   		

\usepackage[table]{xcolor}			
\usepackage{multirow}				
\usepackage{watermark}				
\usepackage{datetime}				
\usepackage{pst-tree}				
\usepackage{algorithm}				
\usepackage{setspace}				
\usepackage{amsmath}					
\usepackage[noend]{algpseudocode}	
\usepackage{multicol}				
\usepackage{tikz}
\usepackage{rotating}
\usepackage{graphicx}
\usepackage{amssymb}
\begin{document}
\frontmatter          
\mainmatter              
\title{Towards Deep Representation Learning with Genetic Programming\thanks{Preprint of paper to appear in EuroGP 2018, LNCS 10781, Springer}}
\titlerunning{Synthesize of Autoencoders}  
%
\author{Lino Rodriguez-Coayahuitl \and
Alicia Morales-Reyes \and
Hugo Jair Escalante}
\authorrunning{Rodriguez-Coayahuitl et al.} 
\institute{Instituto Nacional de Astrofisica, Optica y Electronica, \\
Luis Enrique Erro No.1, Tonantzintla, 72840, Puebla, Mexico,\\
\email{linobi@inaoep.mx}\\ 
}

\maketitle              

\begin{abstract}
Genetic Programming (GP) is an evolutionary algorithm commonly used for machine learning tasks. In this paper we present a method that allows GP to transform the representation of a large-scale machine learning dataset into a more compact representation, by means of processing features from the original representation at individual level. We develop as a proof of concept of this method an \textit{autoencoder}. We tested a preliminary version of our approach in a variety of well-known machine learning image datasets. We speculate that this method, used in an iterative manner, can produce results competitve with state-of-art deep neural networks.

\end{abstract}

\section{Introduction}

Machine learning (ML) algorithms for tasks such as classification, prediction or clustering, require that training samples they are fed with are described in a compact form, in order to deliver an output in a reasonable amount of time. This description is known as \textit{representation}. 

Representation learning is a set of methods that allows a machine to be fed with raw data, such as images pixels, and to automatically discover the representations needed for classification or other machine learning tasks~\cite{lecun2015deep}. In this regard, Deep Learning has showed to be an powerful framework for representation learning.

Deep learning methods consist in different architectures of several stacked layers of artificial neural networks. These deep neural networks (DNN) are based on the idea that a slightly more compact and abstract representation is generated in each forward layer of a DNN; thus with enough of these layers, an original representation consisting in raw data can be transformed into a representation both, compact enough to be tractable by ML algorithms, and abstract enough such that classes or clusters in data can still be discriminated by ML algorithms.

Other methods that can be used for representation learning are matrix factorization~\cite{lee1999learning}, linear discriminant analysis~\cite{mika1999fisher}, principal component analysis~\cite{wold1987principal} and genetic programming (GP)~\cite{koza1992genetic}, among others. However, among all these methods DNN has produced representations that hold the highest records in classification accuracy in many popular machine learning datasets~\cite{lecun2015deep}.

GP is an evolutionary algorithm generally used for machine learning tasks. GP has also been used for representation learning with mixed results~\cite{limon2015class}. GP has achieved competitive results against state-of-the-art methods when the raw initial representation is composed of up to a few dozens variables (known as \textit{features}). When initial representations are composed of hundreds or thousands of features, such as in image or time series problems, GP approaches to representation learning need to leverage from human expert knowledge of the problem's domain in order to achieve competitive results, an undesirable property considering a trend towards higher degrees of automation, and unlike deep learning, which one of its most important aspects is being domain agnostic.

In this work we propose a new method that allows GP to deal with these two shortcomings. Our method is inspired in the same basic idea that motifs DNN, in the sense that we propose a multilayer GP that gradually generates more compact representations. The rest of the paper is as follows. For the rest of this section we describe the basics of GP and then we present related works to our proposed approach. In Sec. 2 we state the representation learning problem and discuss some caveats in attempting to deal with it with GP; in Sec. 3 we present our approach we call Deep Genetic Programming; in Sec. 4 we present some preliminary experimental results; in Sec. 5 we present some concluding thoughts and discuss the future line work.

\subsection{Preliminaries}

\begin{figure}
  \centering
    \includegraphics[width=.40\textwidth]{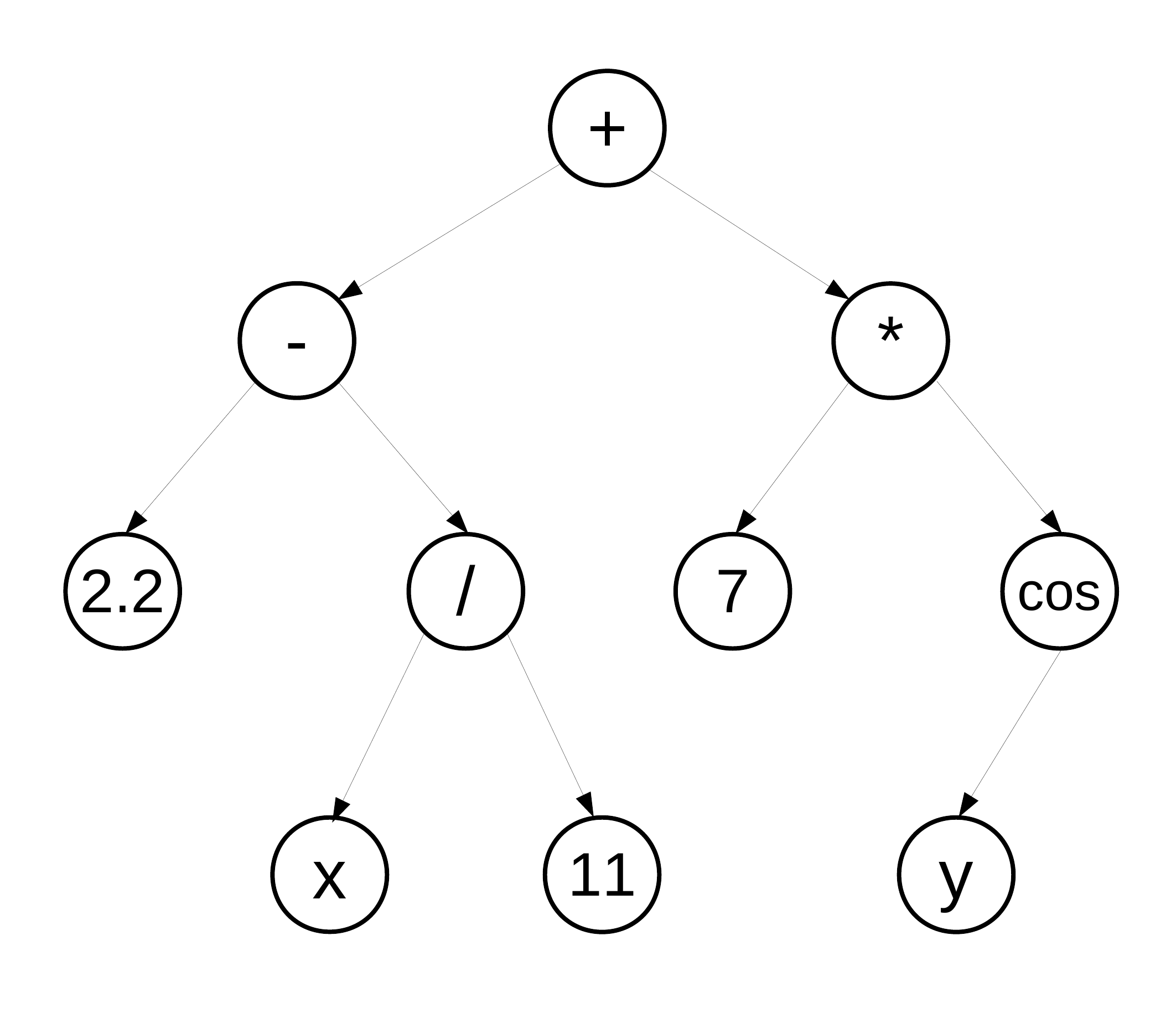}
  \caption[GP individual tree]{Tree structures like this are typically used in GP to represent individual candidate solutions. This instance represents the function $f(x,y) = (2.2 - (\frac{x}{11}))+(7*\cos(y))$.}
  \label{fig:GPTree}
\end{figure}

GP is an evolutionary algorithm whose main difference with other evolutionary algorithms such as genetic algorithms (GA)~\cite{holland1992adaptation} or differential evolution (DE)~\cite{storn1997differential}, is that individuals in GP represent complete mathematical expressions or small computer programs, unlike GA or DE whose individuals are vectors, set of parameters or strings. GP individuals are commonly described by a tree structure. Figure~\ref{fig:GPTree} shows an example of such a structure and the expression it represents. In a GP tree, leaf nodes, called terminals or zero-argument functions, represent input variables or constant values, while inner nodes represent very simple functions, such as arithmetic operands or trigonometric functions. The set of functions from which inner nodes can take their form has to be decided beforehand, and its called the \textit{set of primitives} or simply primitives.

GP individuals are evaluated by an objective function that measures their fitness in the task they are supposed to tackle. Those individuals that perform poorly are discarded from the population and replaced by new individuals generated by performing genetic operations, such as crossover or mutation, on the so-far best performing individuals. This process is repeated a given number of generations, and these transformation and selection processes move the overall population towards promising areas of the search space where individuals perform acceptably.

GP is capable of synthetizing (or discovering) mathematical models than approximate some desired, unknown function, and thus it is potentially on par with other versatile machine learning techniques such as artificial neural networks (ANN).

\subsection{Related work}

Our proposed method for representation learning consist in gradually reduce the dimensionality of the initial representation. We leverage on learning autoencoder-like structures through GP to this end. Autoencoders were, originally, multilayered artificial neural networks (ANN) that attempt to copy its input to its output while data traverses a bottleneck neuron layer formed at the middle of the network. The layers of the network that comprises from its input layer up to the bottleneck it is called "encoder", because its task is to fit the data to pass through the bottleneck layer, with the least amount of information loss possible, while the rest of the network, from the bottleneck up to the output is called "decoder", because its task is to decompress the data from its compressed representaton at the bottleneck layer back to its original form.

Training autoencoders can be a method useful for representation learning~\cite{Goodfellow-et-al-2016}, and it was one of the original methods for training DNN~\cite{hinton2006reducing}. After training an autoencoder with the dataset we are interested in generating a new representation for, we can discard the decoder part, and keep the encoding part that generates a compact representation for the data.

We test our method on image datasets. Examples of representation learning for image processing with GP are \cite{trujillo2006synthesis,shao2014feature}; however these kind of works rely on GP individuals that process images as a whole in each primitive function; primitives might be image addition or substraction, but normally they also tend to be specialized image filters. The GP searches for a combination and workflow of these type of functions that render a new representation useful for classification or detection tasks. The availability of these highly specialized primitives to the GP are a form of human expert knowledge brought to the system by the designer. In contrast, our proposed method process the images at individual pixel level, and only simple arithmetic and generic trigonometric functions form the set of primitives. 

There are other methods for representation learning through GP that are not image processing-specific. Limon et al. \cite{limon2015class} developed a method for representation learning based on GP that utilized arithmetic operations along with few statistics measures. They tested their method on a wide variety of datasets, of different problem domains, and found that GP could learn representations that boosted the performance of a classificator when the input representation consisted in a few dozen features, but beyond that (one hundred or more features) other representation learning methods or the classificator alone performed better. Similarily, Lin et al. \cite{lin2007designing} proposed an approach to binary classification based on generating multiple layers of representations through GP. Although superficially similar to our proposed method, their approach vastly differs from ours: each new representation layer contained a single new feature compared with the previous representation layer from which was generated, the rest of the features are taken directly from such previous layer, whereas our approach aims at generating a representation conformed of completely new feautures in each layer. Their method was in fact developed to tackle problems where initial representatios consist of just a few feautures and adding new features might be beneficial, whereas our method is designed to treat large-scale problems where we actually want to reduce the dimensionality of the initial representation.



\section{Problem statement}

We pose the problem of representation learning as one of dimensionality reduction. From a dataset described in representation $\mathbf{N}$, with an arbitrary number of samples, each sample $s$ represented by feature vector $o_s \in \mathbb{R}^n, \forall s$, we wish to learn a new, more compact and abstract, representation $\mathbf{M}$, such that each sample $s$ is now represented by feature vector $q_s \in \mathbb{R}^m, \forall s$ such that $m \ll n$. 

\subsection{Straightforward GP approach}
\label{straightforward}

One way to attempt to generate representation $\mathbf{M}$ through GP would be as follows: for every feature variable $y_i$ that composes $\mathbf{M}$ we set GP to find a function $f^M_i: \mathbb{R}^n \rightarrow \mathbb{R}$, such that set of input variables of $f^M_i(x_1, x_2, ..., x_n)$ is the set of feature variables that compose the representation $\mathbf{N}$, and that $y_i = f^M_i$.

The problem with this approach is that it poses an untractable search space for GP. Let us suppose, without loss of generality, that each GP tree $t^M_i$ that represents $f^M_i$ is built from $2$-arity primitives, i.e. each $t^M_i$ is a perfect binary tree. Since tree $t^M_i$ could (1) require, conceivably, all $n$ original features to generate $y_i$, (2) $t^M_i$ is a perfect binary tree, and (3) input features can only go in leaf nodes of GP trees, then the height of tree $t^M_i$ is,  approximately at least, $\lceil log_{2}(n) \rceil$, and the number of internal nodes is, approximately, $2^{\lceil log_{2}(n) \rceil}$. For simplicity, let us assume for now on that $n$ is a power of 2, therefore the number of internal nodes of $t^M_i$ is $n$. Now let us suppose that we will use a set of $K$ elegible primitives, then the total size of the search space the GP needs to explore is $\mathcal{O}(mK^n)$. 

This is an optimistic, lower bound estimate, since we are not yet taking into account that constants are probably needed as leaf nodes as well; but then again, this estimate shows us the complexity of the problem we are dealing with.

\begin{example}
Suppose we wish to process a set of images to convert them from a original feature space of 64x64 gray scale pixels to a vector of 32 new features. Hence, $n=4096$ and $m=32$. We are set to search for a GP individual composed of 32 trees; each tree, potentially, of height 12. Suppose we are considering the following set of low level functions $\lbrace +, -, \times, / \rbrace$. The GP needs to search for an optimal individual among, at least, $32 \times 4^{4096}$ distinct possible solutions.
\end{example}

Although evolutionary algorithms are ideally suited to explore search spaces of such exponential growth, we propose that there might exist additional steps to the standard GP that can be taken to improve its efficiency.

\section{Deep Genetic Programming}

The problem of representation learning defines intractable search spaces. Using a standard GP approach to tackle representation learning would be computationally expensive.  Moreover, an important number of poor solutions could be generated and the GP would need to deal with them. 

We propose that considering a structural layered processing of the likes of Deep Learning will allow to significantly improve GP performance to tackle representation learning while reducing the computational burden. The idea is as follows: instead of attempting to build GP trees that convert from representation $\mathbf{N}$ to representation $\mathbf{M}$ in a single step, we generate a series of intermediate representations $\mathbf{L_i}$ that allow to gradually go from representation $\mathbf{N}$ to representation $\mathbf{M}$.

We call this approach \textit{Deep Genetic Programming} (DGP).

\subsection{Construction of Layers in DGP}

Starting form initial representation $\mathbf{N}$, and in order to generate intermediate representation $\mathbf{L_1}$, the set of features that compose $\mathbf{N}$ is partitioned into $c$ small subsets $C_i$, such that $|C_i| \ll |\mathbf{N}|$, $\forall i$. Representation $\mathbf{L_1}$ is also partitioned into $c$ small subsets $K_i$, such that $|C_i| > |K_i|$, $\forall i$.

Each feature $l_{1,j} \in K_i$ is generated by GP tree $t^{L_1}_j$, whose leaf nodes can be feature variables taken only from subset $C_i$ as well as constant values; in other words, $t^{L_1}_j$ represents function $f^{L_1}_j: \mathbb{R}^{|C_i|} \rightarrow \mathbb{R}$, such that $f^{L_1}_j(x_1, x_2, ..., x_w)$, and $(x_1, x_2, ..., x_w) \in C_i$.

Each layer $\mathbf{L_i}$ is built in the same fashion as $\mathbf{L_1}$, relying on the partitioned set of features in $\mathbf{L_{i-1}}$, up until $\mathbf{L_z} = \mathbf{M}$. In this way, the process of dimensionality reduction is done in a gradual manner, unlike in a straightforward single-step approach. 

However, the real key of to succesfully leverage on the proposed layered approach is the partitioning of the layers, as the preliminary results we present in Sec.~\ref{experimental_results} show. In the following section we will also present a simple strategy for the construction of such partitions.

\section{Experimental Results}
\label{experimental_results}

In this section we present some preliminary results of our proposed approach. We implement a 1-layer DGP and compare it against a straightforward GP. The motivation behind these experiments is to serve as a proof of concept of DGP as a well as show the implementation and relevance of its key element, the partitioning of the input representation.

The process of representation learning is performed by the GP itself. The best performing individual that outputs the GP after the last generation serves as a feature extraction engine, that can take as input a sample in its original representation and returns as output the sample in the new representation. Notice that the GP evolves such feature extraction engine and at the same time it also discovers the features it generates as output, i.e. the actual representation learned (because it is not defined beforehand). Therefore the GP attempts to, symbiotically, learn a new representation and the mathematical functions that can generate it.

However, a problem arises when we try to define a way to evaluate the learned, more compact, representation. The answer to this problem is not trivial~\cite{bengio2013representation}. In this work, we tackle the problem through the use of a decoding mechanism that reverses the compact representation to its original form. The average discrepancy between some given dataset in its original form and its the reconstructed version (returned by the decoding mechanism) provides a simple way to evaluate the learned encoded representation, in terms of abridge of information.

Therefore, we set the GP to discover three elements: and encoding mechanism that compacts the input representation; the compact representation itself, i.e. the set of features that conform it; and a decoding mechanism, that provides a supporting role in order to evaluate the learned representation (and the mechanism that generates it). Together, encoder and decoder, form what is known as an \textit{autoencoder}.

\subsection{Individual Representation}

\begin{figure}
  \centering
    \includegraphics[width=.90\textwidth]{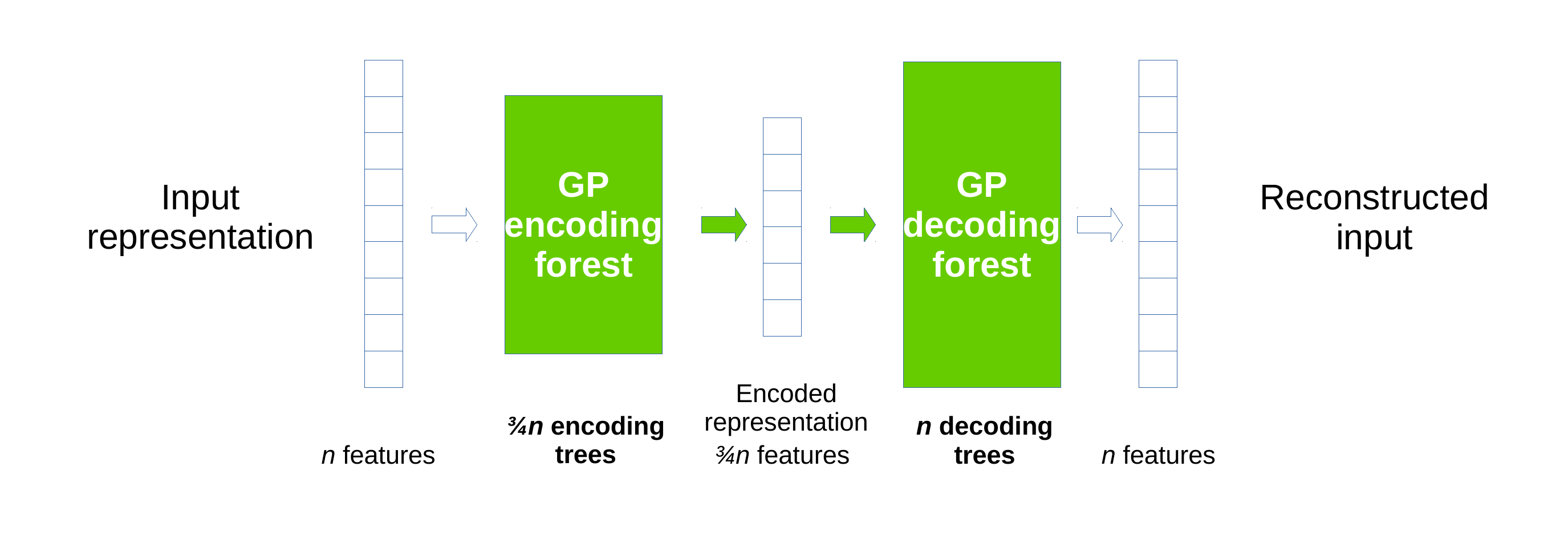}
  \caption[Autoencoder GP individual]{Abstract depiction of a GP autoencoder individual. Each individual consist of two forests connected through a bus the size of the compact representation desired.}
  \label{fig:Individual}
\end{figure}

The individuals design consist in two forests of GP trees connected through a bus of data. One forest is the encoding mechanism, the other forest is the decoding mechanism and the data bus is the new representation for the data. Fig.~\ref{fig:Individual} illustrates this concept.

Given a dataset comprised of an arbitrary number of samples, each one described by the same $n$ features, we wish to reduce the $n$ features to $l_1 = \frac{3}{4}n$ new features, losing the least possible amount of information. That is, from the learned $l_1$ features, it should be possible to reconstruct samples to the original $n$ features.

The encoder will be comprised of a forest of $l_1$ GP trees $t^{L_1}_i$; the decoder will be comprised of a forest of $n$ GP trees $t^{H_0}_i$. Each tree $t^{L_1}_i$ will generate feature $l_{1,i}$ of the compact representation $\mathbf{L_1}$, and each tree $t^{H_0}_i$ will generate feature $h_i$ of reconstructed representation $\mathbf{H_0}$. 

Terminals of each tree $t^{H_0}_i, \forall i$ can only be features of representation $\mathbf{L_1}$ (as well as constant values within some range), i.e. none of these trees can see any of the original features. Similarly, terminals of tree $t^{L_1}_i, \forall i$ can only be taken from the original representation, and they cannot look ahead for features from representation $\mathbf{L_1}$ they are constructing.

As stated before, our GP individual comprehends both encoder and decoder forests integrated as a single indivisible unit we call from now on autoencoder. These autoencoders never get their encoder and decoder components  separated during the evolutionary process, but they exchange small bits of their structure with other autoencoders in a GP population through the evolutionary operators described in the following subsection.

\subsection{Evolutionary parameters and operators}

We will construct a set of randomly generated GP autoencoders individuals, that will constitute an initial population, and through a GP evolutionary process, we will search for an autoencoder that maximizes the average similarity between each sample and its reconstruction, across an entire training dataset. Table~\ref{parameters} shows the set of GP evolutionary parameters used across all experiments performed. 


In each generation, only half of the population is chosen through binary tournament to make it to the next generation. From this half of the population, new individuals are generated with an $.6$ probability of crossover and a $.3$ probability of mutation. Thus, given that population size is fixed to 60 individuals, 30 of them are directly taken from the previous generation, 18 are generated from crossover and 9 are generated through mutation. The remaining 3 individuals are chosen directly from previous generation through elitism.

\begin{table}[]
\centering
\caption{Evolutionary parameters for the GP runs. Arithmetic operands are 2-ary and trigonometric functions are unary primitives. The division function is protected, meaning that any attempt to divide between zero returns as output $1 \times 10^6$, instead of an error. }
\label{parameters}
\begin{tabular}{|l|c|}
\hline
\textbf{Parameter}         & \textbf{Value}                             \\ \hline
Population size   & 60                                \\ \hline
Max. Tree depth   & 4                                 \\ \hline
Set of Primitives & $< +, -, \times, \div, sin, cos>$ \\ \hline
Constants range   & {[}0,1{]}                         \\ \hline
Crossover Prob.   & .6                                \\ \hline
Mutation Prob.    & .3                                \\ \hline
No. Generations   & See Sec.~\ref{setup}              \\ \hline
\end{tabular}
\end{table}

Given that individuals are comprised of two set of trees, special crossover and mutation operations had to be defined that differ from the commonly found in GP literature that deals with single tree GP-individuals. 

In the case of crossover, when two individuals are selected to undergo crossover, randomly selected trees in the encoder forests are exchanged between the two individuals, then the same process is executed again but now for trees in the decoder forests. When an individual is selected for mutation, one randomly selected tree in each its decoder and encoder forests are deleted, these trees are replaced by new randomly generated ones. Notice how none of these operators operate at node level.


\subsection{Objective Function}

To determine similarity between an original sample and the reconstructed output from the autoencoder, we used the mean square error (MSE), defined in Eq.~\ref{eq:MSE}. MSE receives as input original sample $x$ and reconstructed $y$ vectors, and compares them feature by feature, averaging the difference across all of features. MSE output can be thought as a \textit{distance} between a sample and its reconstruction.

\begin{equation}
d_{\mathbf{MSE}}(x,y) = \frac{1}{n} \sum^n_{i=1} (x_i - y_i)^2
\label{eq:MSE}
\end{equation}

\subsection{Experimental setup}
\label{setup}

We implemented and tested three variants of GP algorithms. The first setup consist in a straightfoward GP as described in Sec.~\ref{straightforward}. In this setup, terminals of each tree $t^{L_1}_i, \forall i$ can be any feature from the entire set of $n$ original features. Analogously, terminals of each tree $t^{H_0}_i, \forall i$ can be any of the $l_1$ features from the compact representation $\mathbf{L_1}$. This is our control setup, and its result will serve as a baseline for comparison purposes with the proposed approach.

For the second setup we implemented a 1-layer DGP (1-DGP) along with its partitioning scheme in the following way: we split the initial $n$ input features into $\frac{l_1}{3}$ subsets $C_i$. Associated to each subset $C_i$ there is a subset $K_i$ of features from  $\mathbf{L_1}$, such that each feature $l_{1,j} \in K_i$  is generated by $t^{L_1}_j$. Tree $t^{L_1}_j$ can only \textit{see} the small subset of feautures in $C_i$ 
Four features from input representation $\mathbf{N}$ are assigned to each subset $C_i$.

\begin{figure}
  \centering
    \includegraphics[width=.90\textwidth]{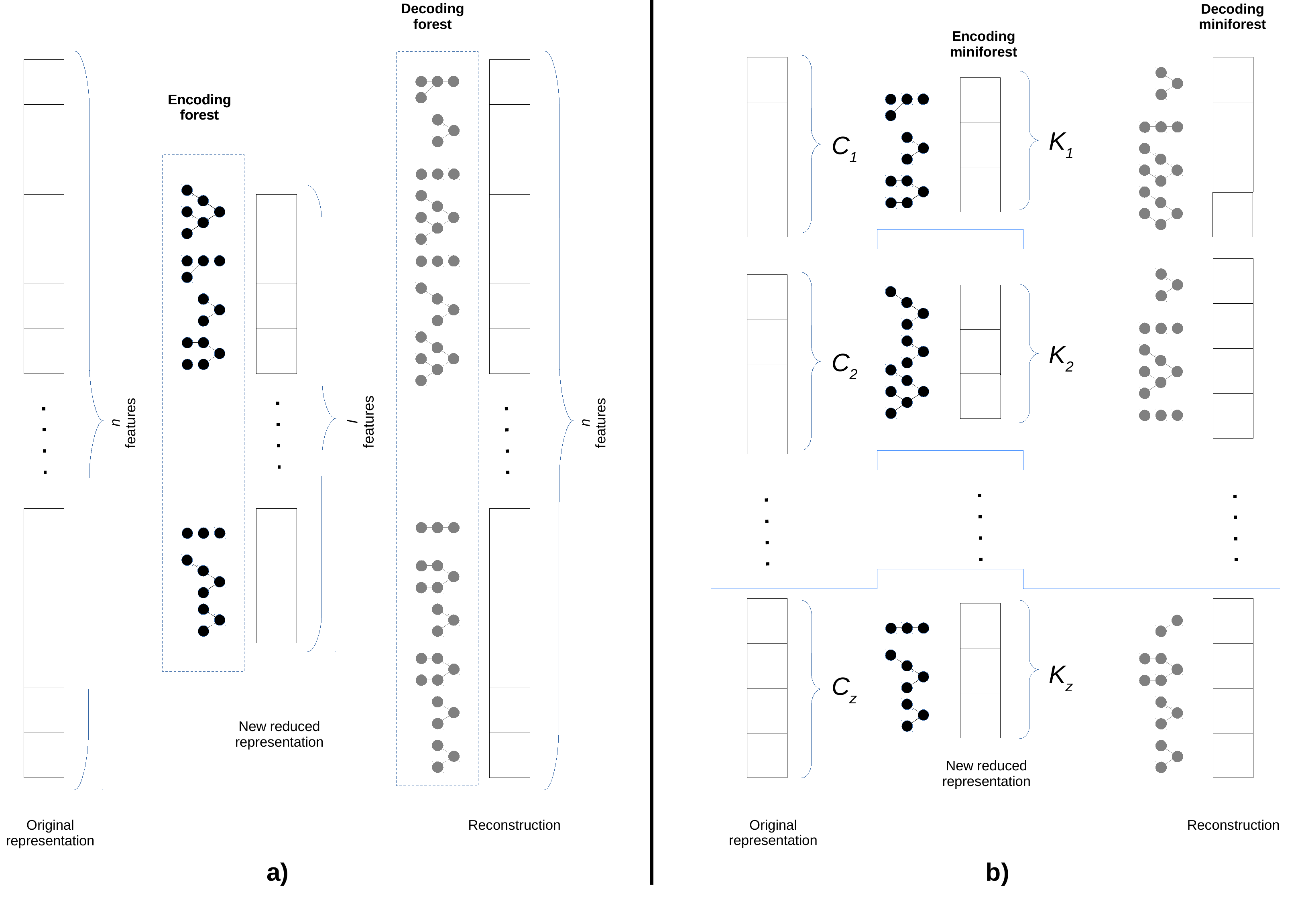}
  \caption[1-Layer Deep]{Comparison between (a) straightforward GP and (b) 1-layer DGP.}
  \label{fig:contrast}
\end{figure}

Mirroring this configuration, the decoding forest is also partitioned in subsets of four trees, such that trees in each subset can only see the subset of three features of some subset $K_i$. All these subsets, (4) input features-(3) encoding trees-(3) new features-(4) decoding trees, are coupled together, to form a miniautoencoder. Fig.~\ref{fig:contrast} contrast this setup against the straightforward GP.

In our third experimental setup we use a \textit{minibatch} based form of training. Instead of presenting the entire dataset to each individual every generation, only a very small, variable, subset of samples (minibatch) of the dataset are presented to each individual in every generation. The number of generations for the GP is chosen such that, \textit{no. of generations $\times$ size of minibatches $=$ size of complete training dataset.} This way we guarantee that the population sees each sample in the training dataset at least once. The minibatches conform a partition, in the mathematical sense, of the entire dataset. The purpose of this setup is to dramatically reduce the computational cost of the GP algorithm, in a similar vein to DNN's usage of stochastic gradient descend.


The objective function in the first two setups described is to minimize the average MSE across all pairs sample-reconstruction from some given dataset. The evolutionary process is executed for 40 generations in both setups. In every generation, each individual of the population is tested against the entire training dataset, the resulting MSEs for every instance in the dataset are averaged, and this result is assigned as the fitness for a given individual. On the other hand, the objective function in the third setup is minimizing avergae MSE for all samples in the current minibatch presented to the population.

We carried all experiments on grayscale image datasets and each pixel is an input feature. We use every four neightboring pixels in the same row to be in subset $C_i$.

\subsection{Results}

We compared the three approaches in ML dataset MNIST~\cite{lecun1998mnist}. For the third setup we also performed a test varying the size of the minibatches and increasing the number of generations in order to allow the GP to see each sample more than just once. After we calibrated the size of minibatches and confirmed that giving more than one pass over training data was beneficial to the proposed method, we further tested it onto two additional ML datasets, namely LFWcrop~\cite{sanderson2014lfwcrop} and Olivetti~\cite{samaria1994parameterisation}. And finally compared the results with those obtained with conventional ANN autoencoders.

Each dataset was split in training and testing set. The evolutionary process is carried with the training set and the top performer individual that results from the process is tested with the testing set, composed of images never before seen during the evolutionary process. Table~\ref{datasets} describes the used datasets.

\begin{table}[]
\centering
\caption{Datasets used for experimentation. All datasets consist in grayscale images; pixel values are normalized to fall in the range [0,1] in all cases.}
\label{datasets}
\begin{tabular}{|l|r|r|r|}
\hline
\textbf{Dataset}                            & \multicolumn{1}{l|}{\textbf{MNIST}} & \multicolumn{1}{l|}{\textbf{LFWcrop}} & \multicolumn{1}{l|}{\textbf{Olivetti}} \\ \hline
Images Resolution (Input features) & 28x28 (784)                & 64x64 (4,096)                & 64x64 (4,096)                 \\ \hline
No. Training samples               & 60,000                     & 12,000                       & 360                           \\ \hline
No. Testing samples                & 10,000                     & 1,233                        & 40                            \\ \hline
\end{tabular}
\end{table}

\begin{figure}
  \centering
    \includegraphics[width=.90\textwidth]{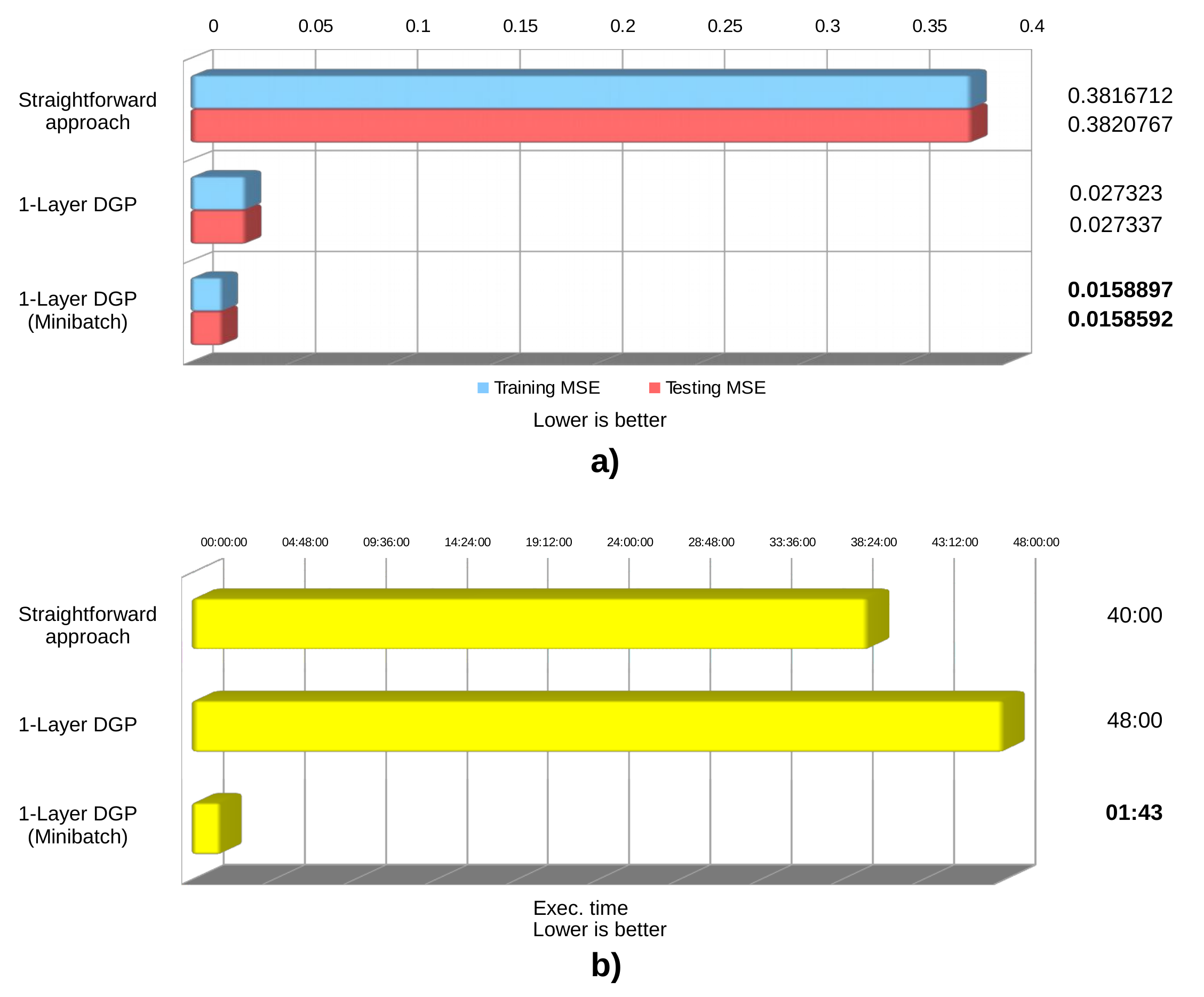}
  \caption[1-Layer Deep]{Results obtained by the three different GP setups (a) MSE across all samples in training and testing datasets. (b) Execution time expressed in hh:mm.}
  \label{fig:Results1}
\end{figure}

Fig.~\ref{fig:Results1} shows the results obtained by the straightforward GP, the 1-Layer DGP and the minibatch training version of it; minibatches were composed of 100 samples. All experiments were done in a workstation with an Intel Xeon CPU with 10 physical cores at 2.9 GHz, with two virtual cores per each physical core, to amount for a total of 20 processing threads, 16 GB of RAM, running Ubuntu Linux 16.04. Algorithms implementation and setups were done by using a in-house software library developed in Python version 3.6. Accelerated NumPy library is used only in the final step of fitness evaluation (averaging the MSE of all sample-reconstruction pairs) of each individual. The straightforward GP can make use of the multiple processing cores by parallelizing the evaluation of sample-reconstructions pairs. On the other hand, both 1-Layer DGP approaches distribute evolution of multiple miniautoencoders across most (but not all) processing threads available, and in this case the evaluations of the sample-reconstructions pairs is done sequentially for each miniautoencoder.

\begin{figure}
  \centering
    \includegraphics[width=.9\textwidth]{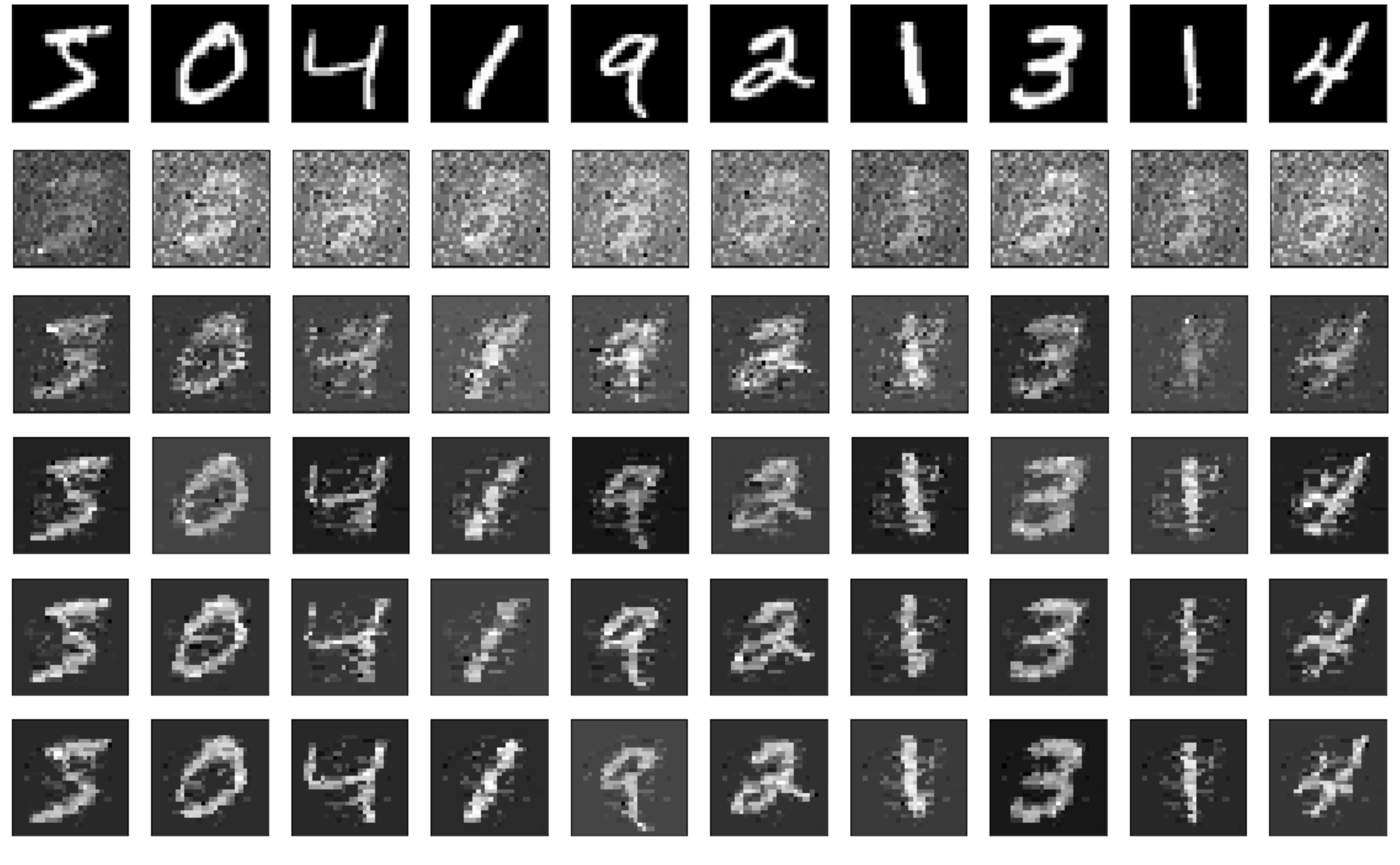}
  \caption[Evolutionary Process Results]{A depiction of the reconstruction generated by the top performer individuals during the evolution process of the 1-Layer DGP. From top row to bottom:the original first ten images of the training set, their best reconstruction obtained after 0, 10, 20, 30 and 40 generations, respectively.}
  \label{fig:Partition_result}
\end{figure}

\begin{figure}
  \centering
    \includegraphics[width=1\textwidth]{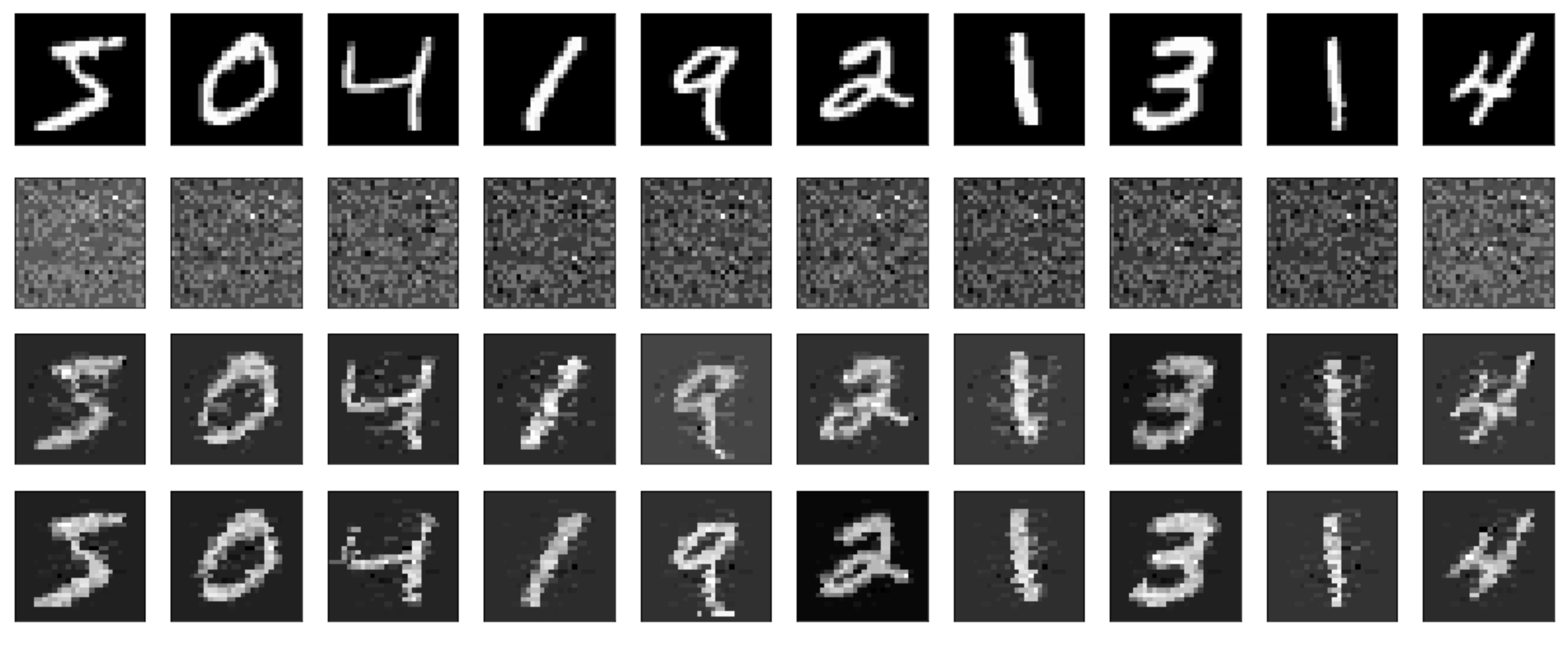}
  \caption[Experimental Results]{Comparisson of the reconstruction of the three experimental setups. From top to bottom: original first 10 images from the training set, best straightforward GP reconstruction, best 1-Layer DGP reconstruction, and best 1-Layer DGP + minibatch training reconstruction.}
  \label{fig:Minibatch_result}
\end{figure}

A visual depiction of the performance of the synthesized autoencoders is shown in Figures \ref{fig:Partition_result} and \ref{fig:Minibatch_result}. Fig.~\ref{fig:Partition_result} shows the gradual increase in performance obtained by the 1-Layer DGP through the evolutionary process. Fig.~\ref{fig:Minibatch_result} compares the reconstruction for the first ten images in the training set, as obtained by the best autoencoders generated by the three different experimental setups. 


Table~\ref{study60000} shows results of varying the size of minibatches to 30, 60, 100, 300, and 600 samples; as well as allowing the algorithm to give one, two and five forward passes over the training data.

\begin{table}[]
\caption[Complete study for full training dataset]{Average MSEs and execution times for a 1-Layer DGP Minibatch approach varying the size of the minibatches to 30, 60, 100, 300, and 600 samples; and giving 1, 2 and 5 forward passes over the sample set.}
\resizebox{\textwidth}{!}{\begin{tabular}{c|ccc|ccc|ccc|ccc|ccc|}
\cline{2-16}
 & \multicolumn{15}{c|}{Mini Batch Size} \\ \cline{2-16} 
 & \multicolumn{3}{c|}{30} & \multicolumn{3}{c|}{60} & \multicolumn{3}{c|}{100} & \multicolumn{3}{c|}{300} & \multicolumn{3}{c|}{600} \\ \hline
\multicolumn{1}{|c|}{Passes} & \multicolumn{1}{c|}{Training} & \multicolumn{1}{c|}{Testing} & \multicolumn{1}{c|}{Time} & \multicolumn{1}{c|}{Training} & \multicolumn{1}{c|}{Testing} & \multicolumn{1}{c|}{Time} & \multicolumn{1}{c|}{Training} & \multicolumn{1}{c|}{Testing} & \multicolumn{1}{c|}{Time} & \multicolumn{1}{c|}{Training} & \multicolumn{1}{c|}{Testing} & \multicolumn{1}{c|}{Time} & \multicolumn{1}{c|}{Training} & \multicolumn{1}{c|}{Testing} & \multicolumn{1}{c|}{Time} \\ \hline

\multicolumn{1}{|c|}{1} & 0.01338 & 0.01334 & 04:44 & 0.01279 & 0.012737 & 03:36 & 0.01345 & 0.013442 & 03:15 & 0.01764 & 0.01764 & 02:49 & 0.02023 & 0.02028 & 02:39 \\ \cline{1-1}

\multicolumn{1}{|c|}{2} & 0.01049 & 0.01045 & 09:31 & 0.01127 & 0.011232 & 07:24 & 0.01153 & 0.011486 & 06:32 & 0.01457 & 0.01449 & 05:30 & 0.01742 & 0.01738 & 05:31 \\ \cline{1-1}

\multicolumn{1}{|c|}{5} & 0.00928 & 0.00921 & 23:27 & \textbf{0.00883} & \textbf{0.008763} & 18:04 & 0.00951 & 0.009502 & 16:27 & 0.01116 & 0.01115 & 14:26 & 0.01418 & 0.01414 & 13:23 \\ \cline{1-1}
\hline
\end{tabular}}
\label{study60000}
\end{table}

\begin{figure}
  \centering
    \includegraphics[width=1\textwidth]{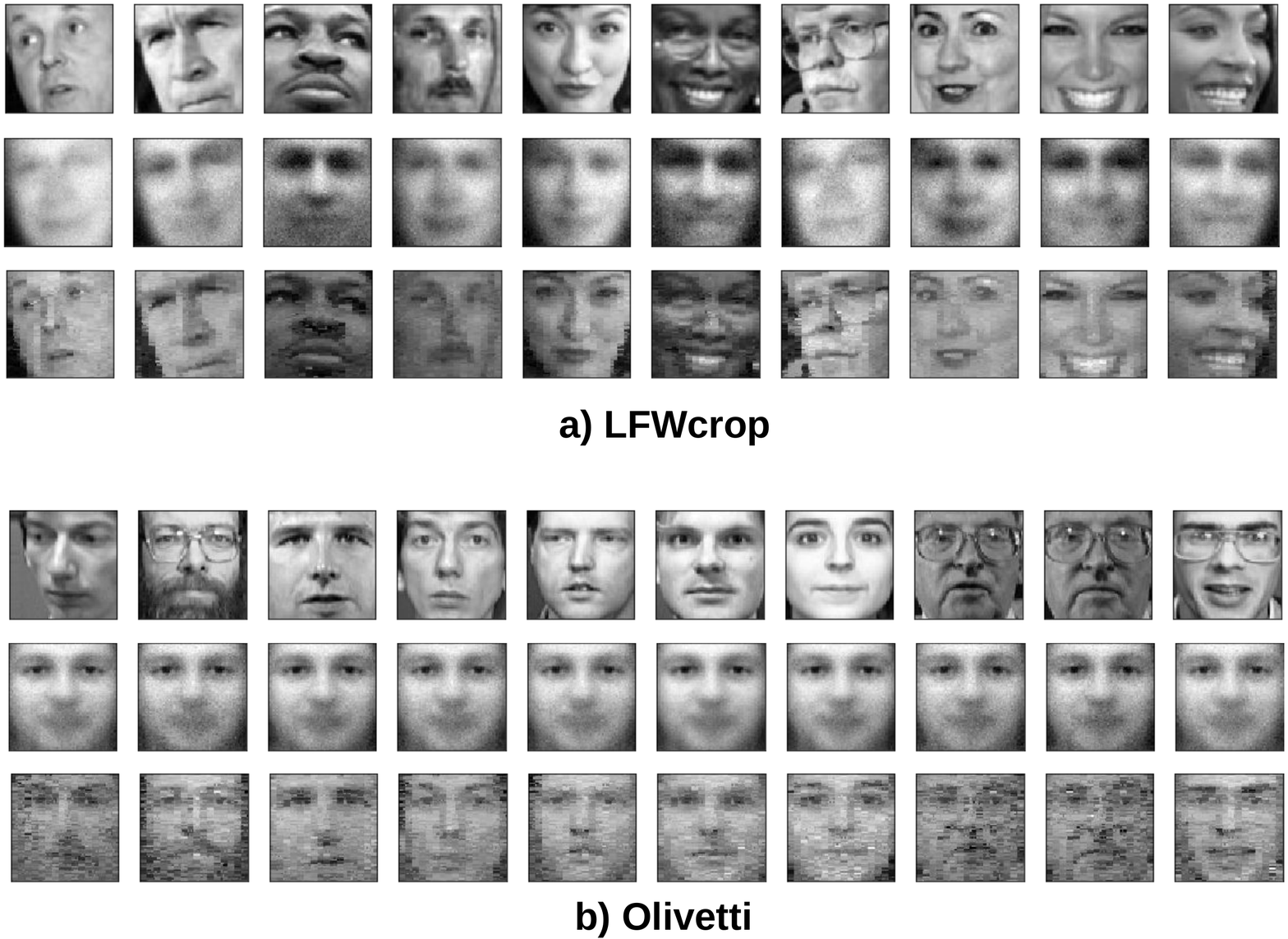}
  \caption[1-DGP vs 1-MLP]{Visual results of the reconstructions generated by the 1-DGP and the 1-MLP in the (a) LFWcrop (b) Olivetti faces datasets. From top to bottom (for both sets) original first ten images from the testing datasets, reconstructions generated by the 1-MLP and reconstructions generated by the 1-DGP}.
  \label{fig:faces}
\end{figure}

We picked up the best performing setup from our minibatch study (minibatches of size 60) and compared its performance against a one hidden layer, fully connected, Multilayer Perceptron (1-MLP) set up as an autoencoder that performs the same compression ratio. We implemented the 1-MLP in TensorFlow Deep Learning library~\cite{abadi2016tensorflow}. We also set the size of the minibatches of the 1-MLP to 60, just as in our GP approach. Table~\ref{comparison} shows the results of each approach, for the different amounts of forward passes/epochs. The main idea behind these experiments is to make a 1-to-1 "layer" and "epoch-to-epoch" comparison in order the study the behavior of the new proposed method and contrast it with conventional methods of deep learning. Fig.~\ref{fig:faces} show a visual appreciation on the reconstructions generated by both autoencoders for LFWcrop and Olivetti samples.

\begin{table}[]
\centering
\caption{MSE results obtained by 1-DGP and 1-MLP autoencoders when tested with testing subsets of different ML datasets, as well as the time required for training/evolution with the training sets of each dataset.}
\label{comparison}
\begin{tabular}{|l|l|c|c|c|c|}
\hline
\multicolumn{1}{|c|}{\multirow{2}{*}{\textbf{Dataset}}} & \multicolumn{1}{c|}{\multirow{2}{*}{\textbf{Passes}}} & \multicolumn{2}{c|}{\textbf{1-DGP}} & \multicolumn{2}{c|}{\textbf{1-MLP}} \\ \cline{3-6} 
\multicolumn{1}{|c|}{}                                  & \multicolumn{1}{c|}{}                                 & Testing              & Time         & Testing              & Time         \\ \hline
\multirow{4}{*}{\textbf{MNIST}}                         & 1                                                     & \textbf{0.01273}     & 03:36:00     & 0.0467               & 00:00:12     \\ \cline{2-6} 
                                                        & 2                                                     & \textbf{0.01123}     & 07:24:00     & 0.0350               & 00:00:24     \\ \cline{2-6} 
                                                        & 5                                                     & \textbf{0.00876}     & 18:04:00     & 0.0215               & 00:01:04     \\ \cline{2-6} 
                                                        & 50                                                    & -                    & -            & \textbf{0.0038}      & 00:11:03     \\ \hline
\multirow{4}{*}{\textbf{Olivetti}}                      & 1                                                     & 0.03232              & 00:08:25     & \textbf{0.0187}      & 00:00:00     \\ \cline{2-6} 
                                                        & 2                                                     & 0.01885              & 00:15:02     & 0.0187               & 00:00:01     \\ \cline{2-6} 
                                                        & 5                                                     & \textbf{0.00948}     & 00:35:23     & 0.0187               & 00:00:03     \\ \cline{2-6} 
                                                        & 50                                                    & -                    & -            & 0.0186               & 00:00:34     \\ \hline
\multirow{4}{*}{\textbf{LFWcrop}}                       & 1                                                     & \textbf{0.00333}     & 04:00:00     & 0.0272               & 00:00:42     \\ \cline{2-6} 
                                                        & 2                                                     & \textbf{0.00258}     & 07:40:37     & 0.0259               & 00:01:10     \\ \cline{2-6} 
                                                        & 5                                                     & \textbf{0.00223}     & 18:30:00     & 0.0245               & 00:03:30     \\ \cline{2-6} 
                                                        & 50                                                    & -                    & -            & 0.0143               & 00:36:23     \\ \hline
\end{tabular}
\end{table}

\subsection{Discussion}

From the results presented in the previous section we can appreciate that the proposed partitioning approach for DGP is one order of magnitude better than a straightforward GP approach. In fact, the straightforward approach does not reach an acceptable solution at all given approximately the same amount of time, making further clear the advange of DGP layer construction process. Even though the difference, in terms of quality of solutions, is not as decisive between the 1-DGP and the its minibatch learning version, the difference in execution time between them is also one order of magnitude. Regarding this last approach, results show that size of minibatches have to be carefully selected in order to get a balance between quality of solution and execution time. We also confirmed that the 1-DGP minibatch technique can benefit from making several passes over the training data.

When compared with a conventional autoencoder, results show that GP beheaves quite different from ANN. GP can quickly (in terms of passes over the training data) build acceptable encoding-decoding models, while an ANN that attempts to generate a representation of 3,072 (588) features from 4,096 (785) initial features, in the case of LFWcrop and Olivetti (MNIST) datasets, has just too many parameters to adjust. This effect is further noticeable as we test datasets with fewer training samples. The GP is simply a more efficient approach in terms of training data usage.

Nevertheless, our GP approach is still behind ANN in terms of total execution time. However this has more to do with the current state of the software implementations of each. While the GP was implemented in pure Python an is prototype grade, TensorFlow is a highly optimized, mature, enterprise grade library. 

\section{Concluding remarks and future work}

In this work we have introduced a new method that allows GP to perform representation learning on large-scale problems without the need of specialized primitives. We provided experimental results that prove the performance gains when compared against a straightforward GP approach.

There is no reason to believe that this method, applied iteratively, can yield representations compact and abstract enough to be usable for other machine learning tasks such as classification or decision taking, and therefore attain results previously though unreachable for GP.

Some of the inmediate questions that arise are:

\begin{itemize}

\item Does subsets $C_i$ have to form a partition? Probably not. In fact, it could be better if some features belong to more than just one subset $C_i$.

\item What other strategies could be used to build subsets $C_i$? So far we have presented a simple strategy based on picking (spatially) neighboring features to be in the same subset $C_i$ but other more complex strategies could exist.

\item How the evolutionary parameters tax on the performance of the algorithm, including the set of primitives available? A complete study is required varying the probabilities of genetic operators, population size, number of available primitives, etc.

\item Why the minibatched version is as good, if not better, in terms of quality than the regular version that sees all training samples each generation? This form of semi-online training probably provides robustness to the population overall, but a more comprehensive answer to this questions is desired.

\end{itemize}

There is still quite a road ahead before we can answer whether or not this method can be competitive with modern state-of-art deep learning methods. The most important question being if adding more layers to a DGP will still keep the upperhand on its side when compared with multilayered ANN. Nevertheless we beleive that many of the necessary ingredients are already there for a GP-based deep learning framework to emerge.

%
%
%
\bibliographystyle{plain}
\bibliography{mybib}{}

\clearpage
\addtocmark[2]{Author Index} 
\end{document}